\documentclass{llncs}

\usepackage{graphicx}
\usepackage{subfigure}
\usepackage{caption}
\usepackage{makeidx}
\usepackage{amsmath,amssymb,theorem,bm,exscale}
\usepackage{textcomp,bm}
\usepackage[latin1]{inputenc}
\usepackage{latexsym}
\usepackage{placeins}
\usepackage{color}
\usepackage{lineno}
\usepackage{array}
\usepackage{multirow}
\usepackage{enumitem}
\usepackage{dsfont}
\usepackage[ruled]{algorithm2e}
\usepackage{float}
\usepackage[pagebackref=false,breaklinks=true,letterpaper=true,colorlinks,urlcolor=blue,citecolor=blue,linkcolor=blue,bookmarks=false]{hyperref}

\begin{document}

\title{Distractor-Aware Neuron Intrinsic Learning for \\ Generic 2D Medical Image Classifications}
\titlerunning{Distractor-Aware Neuron Intrinsic Learning}
\author{Lijun Gong \and Kai Ma \and Yefeng Zheng}
\authorrunning{Gong et al.}
\institute{Tencent Jarvis Lab, Shenzhen, China\\
\email{lijungong@tencent.com}}

\maketitle              
\begin{abstract}
Medical image analysis benefits Computer Aided Diagnosis (CADx).
A fundamental analyzing approach is the classification of medical images, which serves for skin lesion diagnosis, diabetic retinopathy grading, and cancer classification on histological images.
When learning these discriminative classifiers, we observe that the convolutional neural networks (CNNs) are vulnerable to distractor interference.
This is due to the similar sample appearances from different categories (i.e., small inter-class distance).
Existing attempts select distractors from input images by empirically estimating their potential effects to the classifier.
The essences of how these distractors affect CNN classification are not known.
In this paper, we explore distractors from the CNN feature space via proposing a neuron intrinsic learning method.
We formulate a novel distractor-aware loss that encourages large distance between the original image and its distractor in the feature space. The novel loss is combined with the original classification loss to update network parameters by back-propagation.
Neuron intrinsic learning first explores distractors crucial to the deep classifier and then uses them to robustify CNN inherently.
Extensive experiments on medical image benchmark datasets indicate that the proposed method performs favorably against the state-of-the-art approaches.

\end{abstract}
\section{Introduction}
There have been continuous research investigations on medical images in Computer Aided Diagnosis (CADx)~\cite{litjens2017survey} as the automatic identification and analysis of diseases from medical images benefit the clinic diagnosis.
Recently, convolutional neural networks (CNNs) have significantly improved the accuracy of CADx systems and reduced the workload of human screening.
For 2D medical image classification tasks~\cite{wang2020medical}, image classification frameworks (e.g., ResNet~\cite{he2016deep} and EfficientNet~\cite{tan2019efficientnet}) are typically adopted, where a feature extraction backbone~\cite{song2019joint} pre-trained on nature images is applied to get robust low-level features~\cite{song2017stylizing} for the classification network. The collected medical image data is used to finetune the classification network for adaptation~~\cite{deng2009imagenet,litjens2017survey,tajbakhsh2016convolutional}.

In practice, however, we observe that simple yet intuitive finetuning may not achieve favorable results because of the large appearance discrepancy between medical and natural domains.
Moreover, the inter-class appearance difference in medical images is usually smaller than the one in nature images.
As shown in Fig.~\ref{fig:example}, dermatoscopic, fundus and histological images share one thing in common: images in each column appear similar but belong to different categories.
The visual similarity of samples from different categories deteriorates network classification accuracy.
Meanwhile, noisy and blurry effects occur when generating medical images due to limitation of hardware conditions.
Such effects also degrade the image quality for effective classifications.
Therefore, it is desirable to properly handle these aforementioned specific cases, referred as distractors, when training CNNs for better classification accuracy.

\renewcommand{\tabcolsep}{1pt}
\def\swone{0.2\linewidth}
\begin{figure}[t]
\centering
\begin{tabular}{cccc}
    \vspace{-0.5mm}\includegraphics[width=\swone]{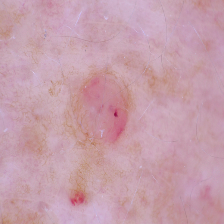}&
    \includegraphics[width=\swone]{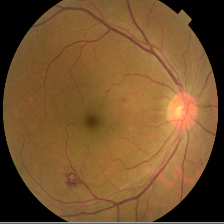}&
    \includegraphics[width=\swone]{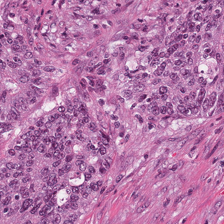}\\
    \vspace{-0.5mm}\includegraphics[width=\swone]{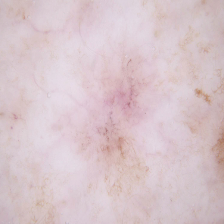}&
    \includegraphics[width=\swone]{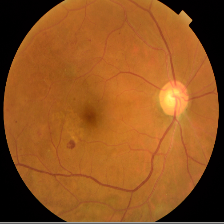}&
    \includegraphics[width=\swone]{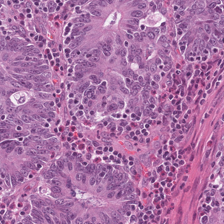}\\
     (a)  & (b)  & (c)  \\
\end{tabular}
\caption{2D medical images. The three columns include (a) dermatoscopic, (b) fundus, and (c) histological images respectively. Each column consists of two similar samples from different categories.}
\label{fig:example}
\end{figure}

In previous works, distractors are mainly selected from input images. For example, an online hard example selection method is proposed in \cite{loshchilov2015online} for image classification. The purpose is to adapt training data to fit the CNN optimizer. Similarly, an online hard example mining (OHEM) is proposed in \cite{shrivastava2016training} to pick up distractors from input images to improve object detection accuracy. It passes all training proposals to the CNN and selects low confidence ones (i.e., hard examples) to update the CNN parameters. More recently, a distractor-aware learning scheme is explored in \cite{zhu2018distractor} to select distractors from input frames for visual tracking. In sum, these methods identify distractors from input images as data augmentation to improve the classification performance. This raises a concern that whether these distractors are selected properly to improve CNNs' performance, especially from the perspective of deep feature space of medical image.

In this paper, we propose a neuron intrinsic learning method to generate distractors in the feature space and then use them to benefit CNN training.
Without modifying CNN structures, we take two rounds of back-propagations during each training iteration. In the first round, we take the partial derivatives of the classification loss with respect to the input image and obtain a response map on the input layer. This response map is named as intrinsic response map (i.e., $A^+$) where each element reflects how much it contributes to the classification loss. Note that in this step, the loss is computed via ground truth labels and the CNN parameters are fixed. Then, we generate another pseudo label, compute the pseudo loss, and generate another intrinsic response map $A^-$ accordingly. This pseudo label indicates the outcome of the distractor effect on the CNN. We trace this label back to the feature space via partial derivatives to generate the distractor (i.e., $A^-$). The elements on $A^-$ show their contribution to produce the pseudo label, which should be kept distant from those on $A^+$.
We organize these response maps with a distraction loss and combine it with the original classification loss.
Using this combined loss, we update the CNN parameters on the second round.
To this end, we explore distractors $A^-$ which are crucial to downgrade deep classifiers (i.e., produce pseudo labels) and use them to improve classification accuracy. We validate the effectiveness of the proposed method on three 2D medical image classification tasks. The results shows that our method performs favorably against state-of-the-art approaches.

\begin{figure}[t]
\includegraphics[width=0.95\textwidth]{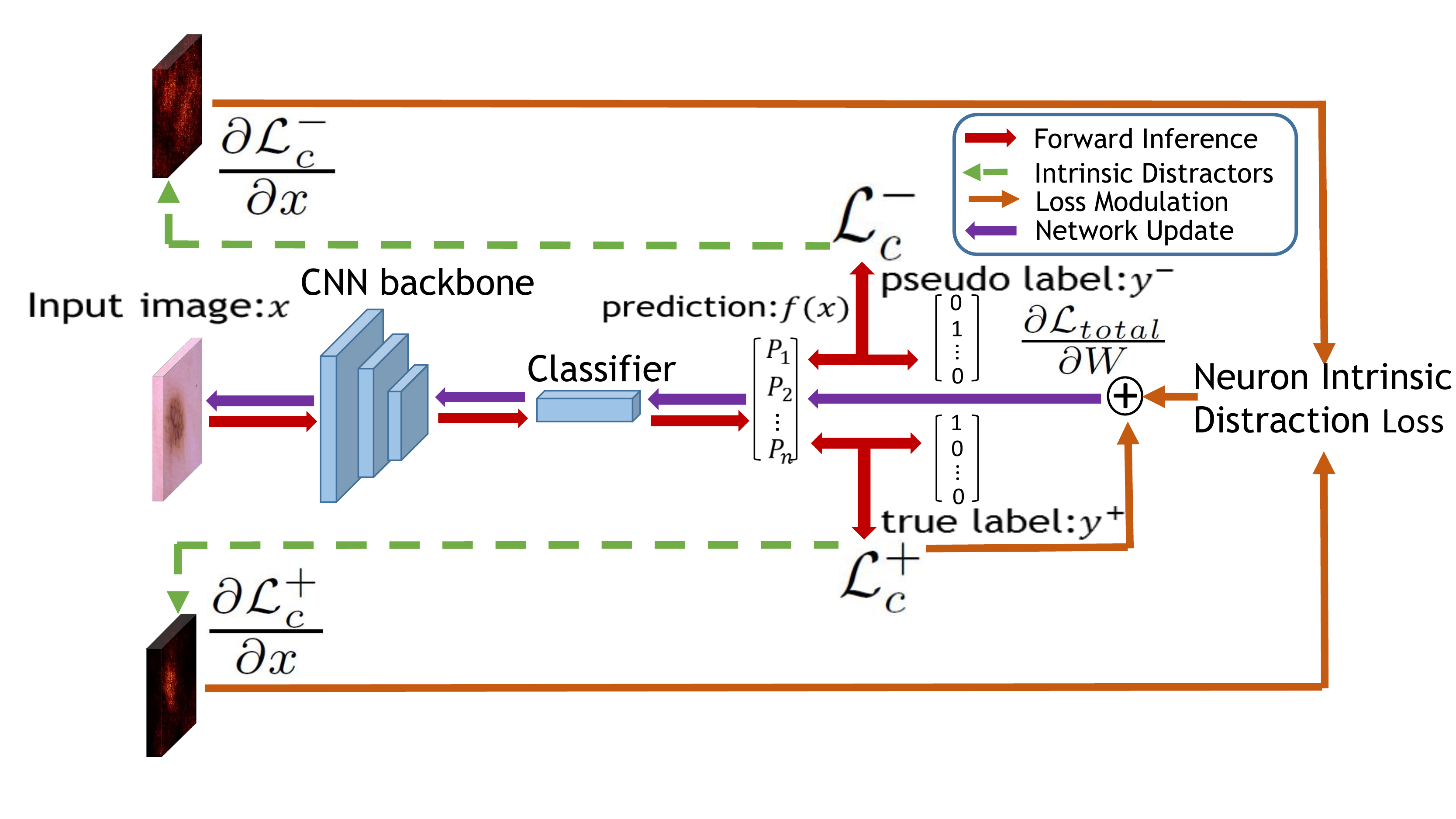}
\centering
\caption{Overview of neuron intrinsic learning for generic medical image classification. During each training iteration, we use two rounds of back-propagations. In the first round, we generate intrinsic response maps by taking $\frac{\partial \mathcal{L}_{c}^+}{\partial {x}}$ and $\frac{\partial \mathcal{L}_{c}^-}{\partial {x}}$. We formulate these maps as a distractor-aware loss term and integrate it into the loss objective function. In the second round, we update the network parameters by taking $\frac{\partial \mathcal{L}_{total}}{\partial W}$.}
\label{fig:pipeline}
\end{figure}

\section{Proposed Method}
Fig.~\ref{fig:pipeline} shows an overview of the proposed method. We do not alter CNN structures while proposing a learning strategy to delve and update CNN parameters. During each training iteration, we use two rounds of back-propagations. In the first round, we generate neuron intrinsic response maps containing a distractor in the CNN feature space.
We formulate a novel distraction loss that encourages large distance between the original image and its distractor in the feature space so as to help increase network's robustness.
In the second round, we use this combined loss to update network parameters. The details are illustrated in the following.

\subsection{Neuron Intrinsic Response Map}
We denote an input image as $x$, the corresponding CNN output as $f(x)$, and loss function as $\mathcal{L}(f(x), y)$. In image classification, $f(x)$ is a vector of scores where each element represents the probability of $x$ belonging to one predefined category, and $y$ is the ground truth label. We can interpret $\mathcal{L}(f(x), y)$ from the Taylor expansion \cite{simonyan2013deep} perspective as:
\begin{equation}\label{eq:intrinsic_map}
   \mathcal{L}(f(x), y) \approx A^\mathsf{T} x + B
\end{equation}
where $A$ is the derivative of the loss $\mathcal{L}(f(x), y)$ and $B$ is a constant value. Eq.~\ref{eq:intrinsic_map} shows that each element in $A$ contributes to $\mathcal{L}(f(x), y)$. Given a specific input sample $x_0$, we can compute $A$ as:
\begin{equation}\label{eq2}
   A = \frac{\partial \mathcal{L}(f(x), y)}{\partial x} \big |_{x=x_{0}}
\end{equation}
We define $A$ as a neuron intrinsic response map as it inherently reflects the response of the network loss. The CNN parameters are fixed when we compute $A$.

\begin{algorithm}[t]
\caption{Distractor-Aware Neuron Intrinsic Learning}
\label{algo}
\LinesNumbered
\KwIn{input image $x$ and ground truth label $y^+$}
$\text{pred} = f(x)$\;
$d_{\text{pred}} = \text{pred.index}(f_m(x)); d = y.\text{index}(y_{\text{m}}^+)$\;
$\mathcal{L}_c^+ = -y^+ [\log (\textup{softmax} (f(x)))]^\mathsf{T}$\;
\eIf{$d_{\text{pred}} \ne d$}{
$y^{-} = [0, 0, ..., 0]$\;
$y^{-}[d_{\text{pred}}] = 1$\;
$\mathcal{L}_c^- = -y^- [\log (\textup{softmax} (f(x)))]^\mathsf{T}$\;
$A^+ = \frac{\partial \mathcal{L}_c^+}{\partial x}; A^{-} = \frac{\partial \mathcal{L}_c^-}{\partial x}$\;
$\mathcal{L}_{d} = \frac{1}{||A^+ - A^{-}||_{2}^{2} + \epsilon}$\;
}{
$\mathcal{L}_{d} = 0$\;
}
$\mathcal{L}_{\text{total}} = \mathcal{L}_c^+ + \lambda \mathcal{L}_{d}$\;
$\mathcal{L}_{\text{total}}.$backward\;
$f.$update\;
\end{algorithm}

\subsection{Distractor Synthesis in the CNN Feature Space}\label{sec:distractors}
Distractors usually confuse CNN to have incorrect predictions. We synthesize a distractor in the CNN feature space based on intrinsic response maps. Given an input image $x$, we denote its ground truth label as $y^+$. This label is an $n$-dimensional vector where there are one element with a value of 1 and remaining elements of 0. The corresponding position of this non-zero element in the vector represents the ground truth category. Besides $y^+$, we define a pseudo label $y^-$ where the non-zero element does not reside in the ground truth category. The classification losses (i.e., softmax and cross-entropy) computed by using $y^+$ and $y^-$ can be written as follows:

\begin{eqnarray}
\label{eq:true-loss}\mathcal{L}_c^+ &=& -y^+ [\log (\textup{softmax} (f(x)))]^\mathsf{T}  \\
\label{eq:pseudo-loss}\mathcal{L}_c^- &=& -y^- [\log (\textup{softmax} (f(x)))]^\mathsf{T}
\end{eqnarray}
We generate two intrinsic response maps $A^+ = \frac{\partial \mathcal{L}_c^+}{\partial x}$ and $A^- = \frac{\partial \mathcal{L}_c^-}{\partial x}$ following Eq. \ref{eq2}. The $A^-$ is defined as the distractor for the current input image. The elements in $A^-$ contribute to the pseudo loss computed in Eq. \ref{eq:pseudo-loss}, which induce the network to make incorrect predictions approaching to $y^-$.
Instead specifying in the image space, we directly synthesize a distractor $A^-$ in the CNN feature space to simulate distractions leading CNN to predict similar scores to pseudo label $y^-$.

Once the distractor is determined, we want to use it to enhance the network's robustness against interference from similar inter-class samples.
An novel distractor-aware objective is set by increasing the distance between $A^+$ and $A^-$. The distraction loss can be written as:
\begin{equation}\label{eq:distraction-loss}
\mathcal{L}_d = \frac{1}{||A^+ - A^-||_{2}^{2} + \epsilon}
\end{equation}
where $||A^+ - A^-||_{2}^{2}$ is the Euclidean distance between $A^+$ and $A^-$, and $\epsilon$ is a small constant value for stable numerical computation. By making $A^+$ and $A^-$ different, the CNN is robust to overcome distractions.

\subsection{Network Training}
We incorporate distractors from Sec. \ref{sec:distractors} during network training. Algorithm \ref{algo} shows the details. During each iteration, we first perform forward propagation to compute the classification loss by using Eq.~\ref{eq:true-loss}, and compute intrinsic response map $A^+$ via a back propagation. Second, we verify if CNN predicts correctly for the current input. We denote the element with maximum value of $f(x)$ as $f_{\text{m}}(x)$.
When the network generates incorrect predictions, we generate $y^-$ by setting  $f_{\text{m}}(x)$ as 1 and the remaining elements as 0 The distraction loss in Eq.~\ref{eq:distraction-loss} can be computed together with the classification loss. The final loss function to train the CNN can be written as:
\begin{equation}
\mathcal{L}_{\text{total}} = \mathcal{L}^+_c + \lambda \mathcal{L}_{d}
\end{equation}
where $\lambda$ is a constant value to balance these two loss terms. We take $\mathcal{L}_{\text{total}}$ to update CNN parameters.

\renewcommand{\tabcolsep}{1pt}
\def\swone{0.21\linewidth}
\begin{figure}[t]
\centering
\begin{tabular}{cccc}
    \vspace{-0.5mm}\includegraphics[width=\swone]{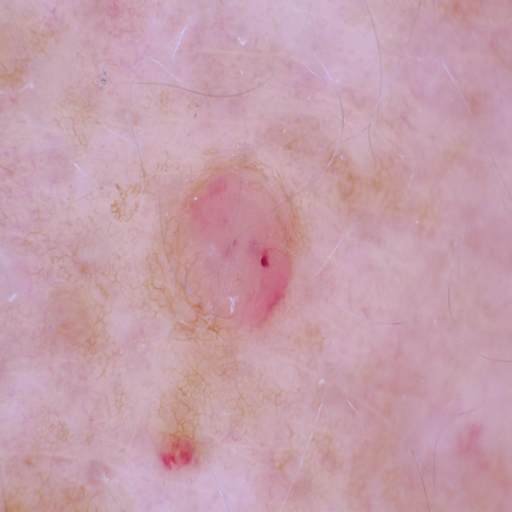}&
    \includegraphics[width=\swone]{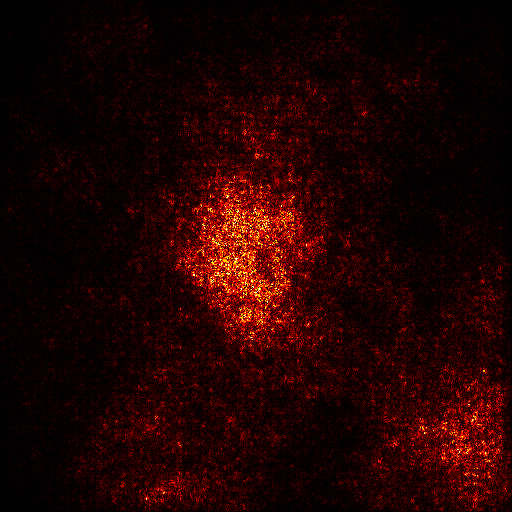}&
    \includegraphics[width=\swone]{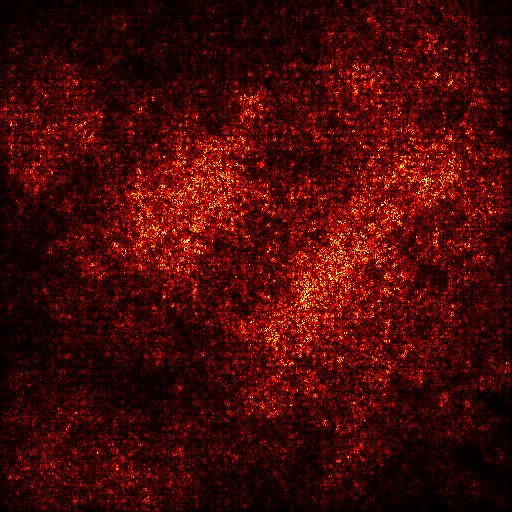}\\
    (a) Dermatoscopic& (b) $A^+$& (c) $A^-$\\
    \vspace{-0.5mm}\includegraphics[width=\swone]{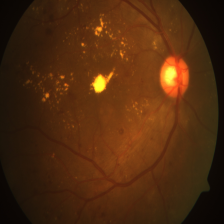}&
    \includegraphics[width=\swone]{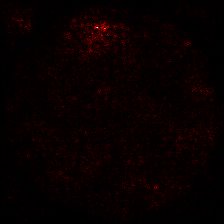}&
    \includegraphics[width=\swone]{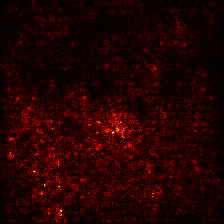}\\
     (d) Fundus & (e) $A^+$ & (f) $A^-$ \\
\end{tabular}
\caption{Visualization of intrinsic response maps. A dermatoscopic image is shown in (a), its corresponding intrinsic response maps are shown in (b) and (c). A fundus image is shown in (d), its corresponding intrinsic response maps are shown in (e) and (f). The pixel values in these maps indicate their contribution to the loss function $\frac{\partial \mathcal{L}_c^+}{\partial x}$ and $\frac{\partial \mathcal{L}_c^-}{\partial x}$, respectively.}
\label{fig:visual}
\end{figure}

{\flushleft \bf Visualization.}
We show that the learned CNN is effective to exclude distractor interference by visualizing the intrinsic response maps shown in Fig.~\ref{fig:visual}. A dermatoscopic image is shown in Fig.~\ref{fig:visual} (a), the corresponding $A^+$ and $A^-$ are shown in Fig.~\ref{fig:visual} (b) and Fig.~\ref{fig:visual} (c), respectively. The high pixel values around the lesion region in (b) indicate that these pixels are extensively utilized to compute $\mathcal{L}_c^+$, while the high pixels values in the background area in Fig.~\ref{fig:visual} (c) indicate these distractors mainly contribute to $\mathcal{L}_c^-$. We enlarge the distance between $A^+$ and $A^-$ to empower the CNN to differentiate distractor interference in the feature space. Another example from a fundus image shows similar performance. Note that the crucial pixels determining fundus classification are around the lension area, not the optic disk area.

\section{Experiments}

We evaluate the proposed method on three 2D medical image classification tasks.
Under each task we compare the proposed method with state-of-the-art approaches on the same benchmark.
These tasks include skin lesion classification, diabetic retinopathy grading, and MSI/MSS classification on histological images.
The corresponding benchmark datasets are HAM10000~\cite{tschandl2018ham10000}, APTOS2019~\cite{APTOS2019}, and CRC-MSI~\cite{histological}, respectively.
In HAM10000, there are 10,015 skin lesion images with predefined 7 categories. We
split images randomly to form a training set and a test set with a ratio of 7:3.
In APTOS2019, there are 3,662 fundus images for grading diabetic retinopathy into five categories. We split the whole dataset randomly into training and test sets where 70\% are for training and 30\% are for testing.
In CRC-MSI, there are 93,408 training and 98,904 testing data for binary classification of histological images. For all the benchmark datasets, we use the averaged F1 score as the evaluation metric.

There are four prevalent CNN backbones (i.e., ResNet50 \cite{he2016deep}, VGGNet11 \cite{simonyan2014very}, InceptionNet V4 \cite{szegedy2016rethinking}, and EfficientNet-b0 \cite{tan2019efficientnet}) utilized during evaluations. The initial weights of all the layers are from the ImageNet pretrained model \cite{deng2009imagenet} except for the last fully connected layer. The learning rate is set as 0.01 and the training iterations are set as 30 epochs. We set $\lambda$ as $1e^{-5}$ and $\epsilon$ as $1e^{-4}$. When comparing with existing methods, we involve deep multi-task learning~\cite{haofu2017deep}, DIL~\cite{rakhlin2018diabetic}, and CANet~\cite{li2019canet}, and report their performance on these datasets.

\subsection{Ablation Study}
The proposed method introduces distractor-aware neuron intrinsic learning into the original classification network. We denote this learning scheme as DANIL. On these three datasets, we show whether performance is improved by integrating DANIL into the baseline network. Table~\ref{Tab:ablation-f1} and~\ref{Tab:ablation-acc} show the evaluation results. The baseline performance (i.e., Base) is consistently improved by using DANIL (i.e., Base + DANIL) on all the benchmark datasets.

\subsection{Comparisons with State-of-the-art}
We compare DANIL with OHEM~\cite{loshchilov2015online,shrivastava2016training} on the benchmark datasets. OHEM is a state-of-the-art hard example mining method that selects distractors from images for CNN training. We denote the original CNN training configuration as Base, the CNN training with OHEM as Base + OHEM, and the CNN training with DANIL as Base + DANIL. Besides, we introduce multi-task prediction~\cite{haofu2017deep}, DIL~\cite{rakhlin2018diabetic} and CANet~\cite{li2019canet} for state-of-the-art comparison.

Table~\ref{Tab:comparison-f1} and~\ref{Tab:comparison-acc} show the evaluation results under averaged F1 and accuracy metrics. We observe that OHEM improves the original CNN classification in the majority of backbones and benchmarks. For some cases (e.g., EfficientNet-b0 on HAM10000), OHEM deteriorates the classification performance under averaged F1 metrics. This indicates that selecting hard examples from images are not robust to improve the CNN performance. In comparison, DANIL is able to consistently make an improvement for different backbones on different datasets. This shows the effectiveness of exploring distractors in the CNN feature space for network training. DANIL performs favorably against existing methods (i.e., multi-task learning, DIL, and CANet) with different CNN backbones as well.

\begin{table}[t]
\centering
\caption{The ablation study using four CNN feature backbones on three benchmarks under averaged F1 metrics.}
\label{Tab:ablation-f1}
\small
\begin{tabular}{p{2.5cm}p{2cm}p{2cm}<{\centering}p{2cm}<{\centering}p{2cm}<{\centering}p{2.2cm}<{\centering}}
\hline
DataSet & Configuration & ResNet50 & VGGNet11 & Inception V4 & EfficientNet-b0\\
\hline
\multirow{2}{*}{HAM10000} & Base & $0.658$ & $0.686$ & $0.671$ & $0.700$ \\
  & Base+DANIL & $\textbf{0.674}$ & $\textbf{0.700}$ & $\textbf{0.690}$ & $\textbf{0.710}$ \\
\hline
\multirow{2}{*}{ATPOS2019} &  Base & $0.617$ & $0.634$ & $0.659$ & $0.642$ \\
 & Base+DANIL & $\textbf{0.660}$ & $\textbf{0.666}$ & $\textbf{0.672}$ & $\textbf{0.671}$ \\
\hline
\multirow{2}{*}{CRC-MSI} & Base & $0.643$ & $0.654$ & $0.650$ & $0.641$\\
 & Base+DANIL & $\textbf{0.654}$ & $\textbf{0.683}$ & $\textbf{0.661}$ & $\textbf{0.652}$ \\
\hline
\end{tabular}
\centering
\caption{The ablation study using four CNN feature backbones on three benchmarks under averaged accuracy metrics.}
\label{Tab:ablation-acc}
\small
\begin{tabular}{p{2.5cm}p{2cm}p{2cm}<{\centering}p{2cm}<{\centering}p{2cm}<{\centering}p{2.2cm}<{\centering}}
\hline
DataSet & Configuration & ResNet50 & VGGNet11 & Inception V4 & EfficientNet-b0\\
\hline
\multirow{2}{*}{HAM10000} & Base & $0.779$ & $0.827$ & $0.830$ & $0.829$ \\
  & Base+DANIL & $\textbf{0.825}$ & $\textbf{0.843}$ & $\textbf{0.845}$ & $\textbf{0.839}$ \\
\hline
\multirow{2}{*}{ATPOS2019} &  Base & $0.804$ & $0.806$ & $0.825$ & $0.801$ \\
 & Base+DANIL & $\textbf{0.825}$ & $\textbf{0.827}$ & $\textbf{0.838}$ & $\textbf{0.831}$ \\
\hline
\multirow{2}{*}{CRC-MSI} & Base & $0.719$ & $0.744$ & $0.728$ & $0.717$\\
 & Base+DANIL & $\textbf{0.735}$ & $\textbf{0.759}$ & $\textbf{0.743}$ & $\textbf{0.732}$ \\
\hline
\end{tabular}
\end{table}

\begin{table}[H]
\centering
\caption{State-of-the-art comparison using four CNN feature backbones on three benchmarks under averaged F1 metrics.}
\label{Tab:comparison-f1}
\small
\begin{tabular}{p{2.5cm}p{2cm}p{2cm}<{\centering}p{2cm}<{\centering}p{2cm}<{\centering}p{2.2cm}<{\centering}}
\hline
DataSet & Method & ResNet50 & VGGNet11 & Inception V4 & EfficientNet-b0\\
\hline
\multirow{4}{*}{HAM10000} & Base & $0.658$ & $0.686$ & $0.671$ & $0.700$ \\
  & Multi-task~\cite{haofu2017deep} & $0.667$ & $0.690$  & $0.679$ & $0.702$ \\
  & Base+OHEM & $0.660$ & $0.692$ & $0.677$ & $0.695$ \\
  & Base+DANIL & $\textbf{0.674}$ & $\textbf{0.700}$ & $\textbf{0.690}$ & $\textbf{0.710}$ \\
\hline
\multirow{5}{*}{ATPOS2019} & Base & $0.617$ & $0.634$ & $0.659$ & $0.642$ \\
 & DIL~\cite{rakhlin2018diabetic} & $0.620$ & $0.637$ & $0.660$ & $0.649$ \\
 & CANet~\cite{li2019canet} & $0.631$ & $0.641$ & $0.664$ & $0.656$ \\
 & Base+OHEM & $0.632$ & $0.644$ & $0.647$ & $0.662$ \\
 & Base+DANIL & $\textbf{0.660}$ & $\textbf{0.666}$ & $\textbf{0.672}$ & $\textbf{0.671}$ \\
\hline
\multirow{3}{*}{CRC-MSI} & Base & $0.643$ & $0.654$ & $0.650$ & $0.641$\\
 & Base+OHEM & $0.649$ & $0.667$ & $0.649$ & $0.644$ \\
 & Base+DANIL & $\textbf{0.654}$ & $\textbf{0.683}$ & $\textbf{0.661}$ & $\textbf{0.652}$ \\
\hline
\end{tabular}
\centering
\caption{State-of-the-art comparison using four CNN feature backbones on three benchmarks under averaged accuracy metrics.}
\label{Tab:comparison-acc}
\small
\begin{tabular}{p{2.5cm}p{2cm}p{2cm}<{\centering}p{2cm}<{\centering}p{2cm}<{\centering}p{2.2cm}<{\centering}}
\hline
DataSet & Method & ResNet50 & VGGNet11 & Inception V4 & EfficientNet-b0\\
\hline
\multirow{4}{*}{HAM10000} & Base & $0.779$ & $0.827$ & $0.830$ & $0.829$ \\
  & Multi-task & $0.811$ & $0.830$  & $0.834$ & $0.828$ \\
  & Base+OHEM & $0.818$ & $0.832$ & $0.830$ & $0.832$ \\
  & Base+DANIL & $\textbf{0.825}$ & $\textbf{0.843}$ & $\textbf{0.845}$ & $\textbf{0.839}$ \\
\hline
\multirow{5}{*}{ATPOS2019} & Base & $0.804$ & $0.806$ & $0.825$ & $0.801$ \\
 & DIL & $0.810$ & $0.806$ & $0.825$ & $0.802$ \\
 & CANet & $0.813$ & $0.810$ & $0.826$ & $0.802$ \\
 & Base+OHEM & $0.813$ & $0.812$ & $0.828$ & $0.814$ \\
 & Base+DANIL & $\textbf{0.825}$ & $\textbf{0.827}$ & $\textbf{0.838}$ & $\textbf{0.831}$ \\
\hline
\multirow{3}{*}{CRC-MSI} & Base & $0.719$ & $0.744$ & $0.728$ & $0.717$\\
 & Base+OHEM & $0.725$ & $0.749$ & $0.732$ & $0.717$ \\
 & Base+DANIL & $\textbf{0.735}$ & $\textbf{0.759}$ & $\textbf{0.743}$ & $\textbf{0.732}$ \\
\hline
\end{tabular}
\end{table}

\section{Concluding Remarks}

Medical image classification was the foundation of automatic computer aided diagnosis. Recent attempts adapted natural image classification models to address this problem.
Their performance was hinged by the discrepancy between medical and natural images. 
In this work, we proposed DANIL to synthesize distractors in the CNN feature space for network learning. We started from the pseudo label, which was the outcome of the distractor interference, and backtraced into the CNN feature space to generate distractors via neuron intrinsic learning. The distractors were kept distant from positive samples in the CNN feature space via the proposed distraction loss. This loss was proposed to learn a more distractor-aware CNN. Extensive experiments on different medical image classification tasks and datasets demonstrated that DANIL improved the CNN classification accuracy and performed favorably against state-of-the-art approaches.
\\
\\
\textbf{Acknowledgments.} This work was funded by the Key Area Research and Development Program of Guangdong Province, China (No. 2018B010111001), National Key Research and Development Project (2018YFC2000702) and Science and Technology Program of Shenzhen, China (No. ZDSYS201802021814180).

%
%
%
\bibliographystyle{splncs04}
\bibliography{ref}

\end{document}